\documentclass[conference]{IEEEtran}
\IEEEoverridecommandlockouts
\usepackage{enumitem}
\usepackage{makecell}

\usepackage{diagbox}
\usepackage{caption}
\usepackage{subcaption}
\usepackage{booktabs}
\usepackage{algorithm}
\usepackage{algpseudocode}
\usepackage{cite}
\usepackage{amsmath,amssymb,amsfonts}
\usepackage{graphicx}
\usepackage{textcomp}
\usepackage{xcolor}
\def\BibTeX{{\rm B\kern-.05em{\sc i\kern-.025em b}\kern-.08em
    T\kern-.1667em\lower.7ex\hbox{E}\kern-.125emX}}
\begin{document}

\title{An Improved Grey Wolf Optimizer Inspired by Advanced Cooperative Predation for UAV Shortest Path Planning\\

\thanks{This work was supported by the XJTLU Research Development Fund (RDF) [grant No. 22-02-096]; XJTLU Teaching Development Fund [grant No. TDF2324-R28-244]; and Guangxi Trusted Software Key Laboratory Project [grant No. KX202314].}
}

\author{
Zuhao Teng\textsuperscript{\dag}, Qian Dong*\textsuperscript{\dag}, Ze Zhang\textsuperscript{\dag},\\
Shuangyao Huang\textsuperscript{\P}, Wenzhang Zhang\textsuperscript{\P}, Jingchen Wang\textsuperscript{\dag},\\
Ji Li\textsuperscript{\ddag}, Xi Chen\textsuperscript{\S} \\
\textsuperscript{\dag}School of Advanced Technology, Xi'an Jiaotong-Liverpool University, Suzhou, China \\
\textsuperscript{\P}School of Internet of Things, Xi'an Jiaotong-Liverpool University, Suzhou, China \\
\textsuperscript{\S}School of Mathematics and Physics, Xi'an Jiaotong-Liverpool University, Suzhou, China \\
\textsuperscript{\ddag}School of Computer Science and Information Security, Guilin University of Electronic Technology, Guilin, China \\
Emails: Zuhao.Teng22@alumni.xjtlu.edu.cn, Qian.Dong@xjtlu.edu.cn
}

\maketitle

\begin{abstract}
With the widespread application of Unmanned Aerial Vehicles (UAVs) in domains like military reconnaissance, emergency rescue, and logistics delivery, efficiently planning the shortest flight path has become a critical challenge. Traditional heuristic-based methods often suffer from the inability to escape from local optima, which limits their effectiveness in finding the shortest path. To address these issues, a novel Improved Grey Wolf Optimizer (IGWO) is presented in this study. The proposed IGWO incorporates an Advanced Cooperative Predation (ACP) and a Lens Opposition-based Learning Strategy (LOBL) in order to improve the optimization capability of the method. Simulation results show that IGWO ranks first in optimization performance on benchmark functions F1–F5, F7, and F9–F12, outperforming all other compared algorithms. Subsequently, IGWO is applied to UAV shortest path planning in various obstacle-laden environments. Simulation results show that the paths planned by IGWO are, on average, shorter than those planned by GWO, PSO, and WOA by 1.70m, 1.68m, and 2.00m, respectively, across four different maps.
\end{abstract}


\section{Introduction}
Over the past few years, the implementation for UAVs has continued to grow across domains including military reconnaissance, environmental sensing, and logistics due to their flexibility, low cost, and rapid deployment \cite{hu2024multi, hu2025computation}. As their applications expand, there is a growing demand for autonomous and reliable operation in complex environments. Path planning, a core component of autonomy, aims to compute an optimal trajectory from start to destination while avoiding obstacles and minimizing time or energy consumption. Its effectiveness is crucial to both mission success and flight safety \cite{maboudi2023review}.

In recent years, numerous studies have emerged focusing on UAV path planning. Existing techniques are generally grouped into three main types: graph-based methods \cite{prasad20223, 9165710, guo2022feedback, huang2024density}, Artificial Intelligence (AI)-based methods\cite{dhuheir2022deep, xi2024lightweight, 9165710, QU2020106099}, and heuristic-based methods \cite{phung2021safety, yu2023hybrid, yu2022novel}.

Graph-based methods model the flight space as a grid or graph, treating path planning as a shortest path search. Though effective, they become computationally expensive in large or complex environments with irregular obstacles or added constraints, limiting real-world applicability. To address this, Prasad et al. \cite{prasad20223} integrated Dijkstra’s algorithm with dynamic threat modeling for safer UAV navigation. Luo et al. \cite{9165710} incorporated probabilistic threat levels into edge weights for risk-aware planning. Guo et al. \cite{guo2022feedback} proposed Feedback RRT*, enhancing planning efficiency through risk-based feedback expansion.

AI-based approaches demonstrate strong capability in handling high-dimensional and large-scale path planning problems \cite{dhuheir2022deep, xi2024lightweight, bayerlein2021multi}. However, they often require extensive training data and struggle to generalize across different environments. Dhuheir et al. \cite{dhuheir2022deep} proposed a DRL-based framework that enables joint trajectory optimization and decentralized CNN processing within UAV swarms to minimize latency. Xi et al. \cite{xi2024lightweight} developed ASAC, a lightweight RL-based algorithm that enhances robustness and efficiency during dynamic route generation in dense urban environments.

Heuristic-based methods formulate path planning as an optimization problem and provide efficient near-optimal results for Non-deterministic Polynomial-time hard (NP-hard) problems without the training overhead of AI-based approaches \cite{chen2021clustering}. Meanwhile, unlike graph-based methods, they scale well to large and high-dimensional search spaces, making them particularly suitable for complex UAV path planning tasks. In addition, they are adaptable to dynamic environments and computationally efficient, which further supports their practical use in real-time UAV applications. However, existing heuristic algorithms often suffer from limited fine-tuning capability and are prone to getting trapped in local optima \cite{rajwar2023exhaustive}. To improve the optimization capability and obtain better path planning solutions, this study proposes a novel IGWO applied to UAV shortest path planning. The principal contributions of this study are summarized as:
\begin{enumerate}[leftmargin=0pt]
\item A novel optimization strategy, ACP, is proposed. It incorporates the information-sharing behavior observed in grey wolf hunting and fully leverages the leadership of the leading wolf, enabling search agents to perform more refined exploration in its vicinity. This mechanism enhances the algorithm’s ability to locate higher-quality solutions.
\item A novel heuristic algorithm, IGWO, is proposed. Specifically, ACP is incorporated prior to the exploitation phase of the traditional Grey Wolf Optimizer (GWO), and the LOBL strategy is integrated after the exploitation phase of GWO. These two strategies improve the method’s performance across fine-level search and avoiding local optima.
\item An IGWO-based method targeting UAV shortest route design has been introduced in this study. Specifically, the objective function for the UAV shortest path planning problem is designed, and IGWO is employed to search for its minimum. Simulation results show that the paths planned by IGWO are, on average, shorter than those planned by GWO, PSO, and WOA by 1.70m, 1.68m, and 2.00m, respectively, across four different maps.
\end{enumerate}

\section{Methodology}
\subsection{Implementation of Improved Grey Wolf Optimizer}
\subsubsection{Grey Wolf Optimizer}
IGWO is designed as an enhanced version of the GWO \cite{mirjalili2014grey}, which operates through two primary phases: exploration and exploitation. Agent exploration dynamics throughout the exploration phase are determined by the following equations:
\begin{equation}
\vec{A} = 2\vec{a} \cdot \vec{r}_1 - \vec{a}
\end{equation}
\begin{equation}
\vec{D} = \left| 2 \cdot \vec{r}_2 \cdot \vec{X}_p(t) - \vec{X}(t) \right|
\end{equation}
\begin{equation}
\vec{X}(t+1) = \vec{X}_p(t) - \vec{A} \cdot \vec{D}
\end{equation}

Where the term $\vec{X}_p(t)$ denotes the prey’s position at the current iteration. The positions of the search agents at time steps $t$ and $t+1$ are denoted by $\vec{X}(t)$ and $\vec{X}(t+1)$, respectively. The coefficients $\vec{r_1}$ and $\vec{r_2}$ are uniformly distributed random values within the interval $(0, 1)$, while $\vec{a}$ is a control parameter that linearly declines from 2 to 0 throughout the iterations.

The exploitation phase of the GWO can be expressed by the following equations:
\begin{equation}
\scalebox{0.8}{$
\begin{aligned}
\vec{D}_{\alpha} &= \left| 2 \cdot \vec{r}_2 \cdot \vec{X}_{\alpha}(t) - \vec{X}(t) \right|, \\
\vec{D}_{\beta}  &= \left| 2 \cdot \vec{r}_2 \cdot \vec{X}_{\beta}(t) - \vec{X}(t) \right|, \quad
\vec{D}_{\delta} &= \left| 2 \cdot \vec{r}_2 \cdot \vec{X}_{\delta}(t) - \vec{X}(t) \right|
\end{aligned}
$}
\end{equation}

\begin{equation}
\begin{aligned}
\vec{X}_{1}(t) &= \vec{X}_{\alpha}(t) - \vec{A} \cdot \vec{D}_{\alpha}, \\
\vec{X}_{2}(t) &= \vec{X}_{\beta}(t) - \vec{A} \cdot \vec{D}_{\beta}, \quad
\vec{X}_{3}(t) = \vec{X}_{\delta}(t) - \vec{A} \cdot \vec{D}_{\delta}
\end{aligned}
\end{equation}
\begin{equation}
\vec{X}(t+1) = \frac{\vec{X}_{1}(t) + \vec{X}_{2}(t) + \vec{X}_{3}(t)}{3}
\end{equation}

Where the positions of the top three leading search agents at time step $t$ are denoted by $\vec{X}_{\alpha}(t)$, $\vec{X}_{\beta}(t)$, and $\vec{X}_{\delta}(t)$, and $\vec{X}(t)$ represents any individual search agent's position. The updated position for the next iteration is expressed as $\vec{X}(t+1)$.

\subsubsection{Advanced Cooperative Predation}
Grey wolves are social predators whose hunting behavior involves not only strength and speed, but also a high degree of coordination \cite{muro2011wolf}. The underlying coordination arises from information sharing among individuals, which is abstracted in this work through the use of the population centroid to model collective motion dynamics. The population centroid is calculated as:
\begin{equation}
\vec{X}_{mean}(t) = \frac{1}{n} \sum_{i=1}^{n} \vec{X}_{i}(t)
\end{equation}

Where $n$ indicates the total count of search agents, $\vec{X}_{i}(t)$ specifies the location associated with the $i$-th agent during iteration $t$, while $\vec{X}_{\text{mean}}(t)$ denotes the average spatial coordinate derived from all individuals in the group.

Grey wolves typically engage in encircling behavior when tracking their prey. As they approach the target, they exhibit a dynamic spiral movement pattern influenced by both the population's centroid position and the leading wolf's position. This dynamic is captured through the equations below:
\begin{equation}
\scalebox{0.8}{$
\vec{X}(t+1) = 
\begin{cases}
r_3 \cdot \vec{X}_{\text{mean}}(t) + \gamma \cdot (\vec{X}_{\alpha}(t) - \vec{X}(t)), & \text{if } r_3 < 0.5 \\
r_3 \cdot \vec{X}_{\text{mean}}(t) + \gamma \cdot \left( \frac{\vec{X}_{\beta}(t) + \vec{X}_{\delta}(t)}{2} - \vec{X}(t) \right), & \text{otherwise}
\end{cases}
$}
\end{equation}
\begin{equation}
\gamma = 2e^{r_4^{\frac{T - t + 1}{T}}} \cdot \sin(2\pi r_4)
\end{equation}

Where $T$ denotes the maximum iteration count, $t$ indicates the current loop number, and $r_3$ and $r_4$ are uniformly sampled random variables within the interval $(0, 1)$.

In Eq. (9), $\gamma$ is the spiral factor, which guides the wolves to update their positions in a spiral pattern, thereby facilitating fine-grained exploration. In Eq. (8), when $r_3 < 0.5$, the search agents adjust their locations by incorporating the population centroid and the location of the top-ranked $\alpha$ wolf. Conversely, when $r_3 \geq 0.5$, the agents follow the centroid and the average position of the second- and third-ranked $\beta$ and $\delta$ wolves. These two scenarios occur with equal probability, indicating that the agents make full use of both the population centroid and the positional information of the top three wolves. This strategy enhances the algorithm’s ability to perform a refined search around promising areas, thus increasing the likelihood of obtaining high-quality solutions.

\subsubsection{Lens Opposition-based Learning}
LOBL, originally proposed by Tizhoosh \cite{tizhoosh2005opposition}, is a strategy aimed at improving the optimization ability of heuristic optimization methods. It utilizes a concept from convex lens optics to generate a mirrored counterpart for each agent with respect to a central axis. This reflective transformation increases the algorithm’s capacity to avoid stagnation in suboptimal regions. In this context, agents are updated using the expression given below:
\begin{equation}
\vec{X}(t+1) = \frac{a_j + b_j}{2} + \frac{a_j + b_j}{2k} - \frac{\vec{X}(t)}{k}
\end{equation}

Where factor $k = \frac{h}{h^*}$ quantifies the magnification induced by the convex lens effect, describing how the agent’s mirrored position is scaled relative to its original offset from the focal axis. $\vec{X}(t+1)$ and $\vec{X}(t)$ denote the positions of the search agent at the next and current iterations, respectively. $a_j$ and $b_j$ indicate the interval endpoints that constrain the agent's movement.

\subsubsection{Improved Grey Wolf Optimizer}
IGWO is developed from the original GWO with two strategies: ACP and LOBL. In the conventional GWO, the information provided by the three elite individuals is not fully exploited, which limits the algorithm’s fine-tuning capability. ACP is introduced to address this issue. It is incorporated before the exploitation phase of GWO, thereby enhancing information sharing between search agents and elite individuals. As a result, more agents are guided to conduct refined searches in the vicinity of the elite solutions.

\begin{figure}[!t]
    \centering
    \includegraphics[width=0.98\columnwidth]{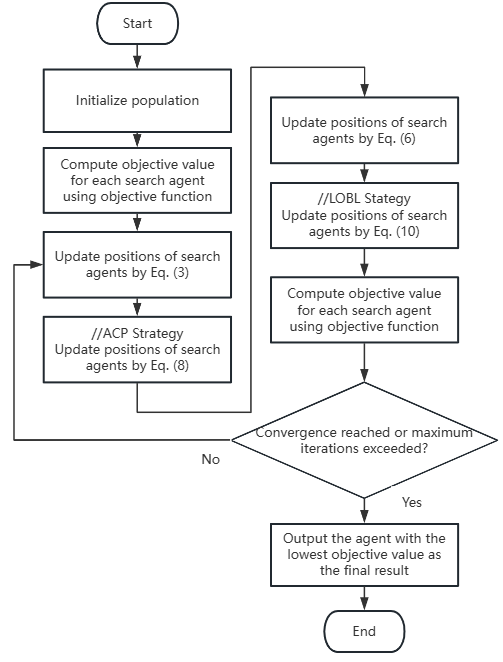}
    \caption{Flowchart of IGWO}
    \label{flowchart}
\end{figure}

One common limitation of heuristic algorithms, including GWO, lies in their tendency to prematurely converge, which can trap the search process in suboptimal regions and reduce overall exploration effectiveness. To mitigate this issue, the LOBL strategy is integrated into IGWO. Specifically, LOBL is applied after the exploitation phase of the standard GWO to promote solution diversity and prevent convergence to inferior regions. The flowchart of the IGWO is depicted in Fig. \ref{flowchart}.

\begin{table*}[!htbp]
\centering
\scriptsize  
\setlength{\tabcolsep}{2.5pt}  
\caption{Statistical comparison of IGWO and three other algorithms on benchmark functions}
\label{IGWO_table}
\begin{tabular}{@{}lccccccccccccccc@{}}
\toprule
 & \multicolumn{2}{c}{F1} & \multicolumn{2}{c}{F2} & \multicolumn{2}{c}{F3} & \multicolumn{2}{c}{F4} & \multicolumn{2}{c}{F5} & \multicolumn{2}{c}{F6} & \multicolumn{2}{c}{F7} \\
 & Avg & Std & Avg & Std & Avg & Std & Avg & Std & Avg & Std & Avg & Std & Avg & Std \\
\midrule
IGWO & 1.94E-76 & 8.96E-76 & 2.45E-20 & 1.23E-39 & 1.91E-62 & 1.05E-61 & 1.41E-38 & 4.44E-38 & 2.73E+1 & 7.20E-1 & 8.67E-1 & 3.33E-1 & 2.90E-4 & 2.53E-4 \\
PSO  & 6.80E-3 & 1.86E-2 & 7.9E-2 & 8.30E-2 & 3.89E+2 & 1.95E+2 & 3.80E+0 & 1.61E+0 & 9.28E+1 & 1.11E+2 & 4.38E-3 & 9.43E-3 & 2.88E-2 & 1.33E-2 \\
GWO  & 3.07E-10 & 2.73E-10 & 1.06E-06 & 3.94E-07 & 1.30E+0 & 1.54E+0 & 1.80E-2 & 1.08E-2 & 2.75E+1 & 7.07E-1 & 1.07E+0 & 3.82E-1 & 4.23E-3 & 1.71E-3 \\
WOA  & 6.03E-29 & 2.49E-28 & 1.07E-20 & 3.07E-20 & 6.45E+4 & 1.72E+4 & 5.69E+1 & 2.69E+1 & 2.86E+1 & 2.69E-1 & 7.69E-1 & 2.40E-1 & 9.98E-3 & 1.46E-2 \\
\midrule
 & \multicolumn{2}{c}{F8} & \multicolumn{2}{c}{F9} & \multicolumn{2}{c}{F10} & \multicolumn{2}{c}{F11} & \multicolumn{2}{c}{F12} & \multicolumn{2}{c}{F13} \\
 & Avg & Std & Avg & Std & Avg & Std & Avg & Std & Avg & Std & Avg & Std \\
\midrule
IGWO & -1.63E+6 & 1.87E+6 & 0 & 0 & 4.44E-6 & 0 & 0 & 0 & 4.29E-2 & 2.05E-2 & 7.22E-1 & 2.67E-1 \\
PSO  & -6.84E-3 & 8.04E+2 & 4.62E+1 & 1.11E+1 & 9.62E-1 & 8.2E-1 & 2.14E-2 & 2.25E-2 & 1.79E-1 & 2.67E-1 & 2.81E-1 & 7.58E-1 \\
GWO  & -5.68E+3 & 1.05E+1 & 1.05E+1 & 5.96E+0 & 4.12E-6 & 1.92E-6 & 8.63E-3 & 1.41E-2 & 5.48E-2 & 2.45E-2 & 8.66E-1 & 2.12E-1 \\
WOA  & -9.68E+3 & 1.64E+3 & 3.31E-2 & 1.82E-1 & 1.02E-14 & 5.35E-15 & 3.65E-2 & 1.40E-1 & 6.18E-2 & 9.48E-2 & 7.52E-1 & 3.16E-1 \\
\bottomrule
\end{tabular}
\end{table*}

\subsection{IGWO for UAV Shortest Path Planning}
In this section, the UAV path is abstracted as a sequence of discrete points. The optimal path is defined as the one that minimizes the total distance between all consecutive points while avoiding any intersection with obstacles. Heuristic methods formulate it as a search for the optimal solution within a defined decision space. This formulation leads to the construction of the following mathematical objective:
\begin{equation}
\scalebox{0.8}{$
\arg\min_{\mathbf{X}} f(\mathbf{X}) =
\begin{cases}
\sum\limits_{i=0}^{m-2} \sqrt{(x_{i+1} - x_i)^2 + (y_{i+1} - y_i)^2}, & \text{if } nO = 0 \\
P \times nO, & \text{if } nO > 0
\end{cases}
$}
\end{equation}

Where $(x_i, y_i)$ and $(x_{i+1}, y_{i+1})$ represent the coordinates of the $i$-th and $(i+1)$-th points on the path, respectively. The total number of points along the path is represented by $m$, including both the starting and ending points. $P$ is a penalty coefficient, typically set to 10. $\mathbf{X} = (x_1, y_1, x_2, y_2, \ldots, x_{m}, y_{m})$ denotes the vector that concatenates the coordinates of all points along the path. $nO$ denotes the number of times the path intersects with obstacles. Notably, when $nO > 0$, a large penalty is imposed on the objective function to ensure that the resulting path avoids obstacle collisions. Conversely, when $nO = 0$, the objective is reduced to minimizing the Euclidean distance between consecutive points along the path, thereby yielding the shortest path.

When solving the shortest path problem using the IGWO, an initial population of 40 search agents is generated, with each agent representing a candidate path. These 40 agents are evaluated using Eq. (11) to obtain their corresponding objective function values. Subsequently, the positions of the agents are iteratively updated according to Fig. \ref{flowchart}. After each update, all agents are re-evaluated using Eq. (11) to compute their current objective values. When either a convergence threshold is satisfied or the algorithm completes its iteration cycle, the search agent with the lowest objective function value is selected as the final solution.

During the iterative process, the feasibility of each path is continuously evaluated based on its collision status with predefined obstacles. Infeasible paths, which intersect with obstacles, are penalized via the $P \times nO$ term in the objective function. This penalty-based approach ensures that the algorithm is discouraged from exploring trajectories that intersect with obstacles. Meanwhile, feasible paths are assessed based on their total Euclidean distance, promoting the discovery of shorter solutions.

\begin{table}[!t]
    \centering
    \caption{Characteristics of Benchmark Functions} 
    \includegraphics[width=\columnwidth]{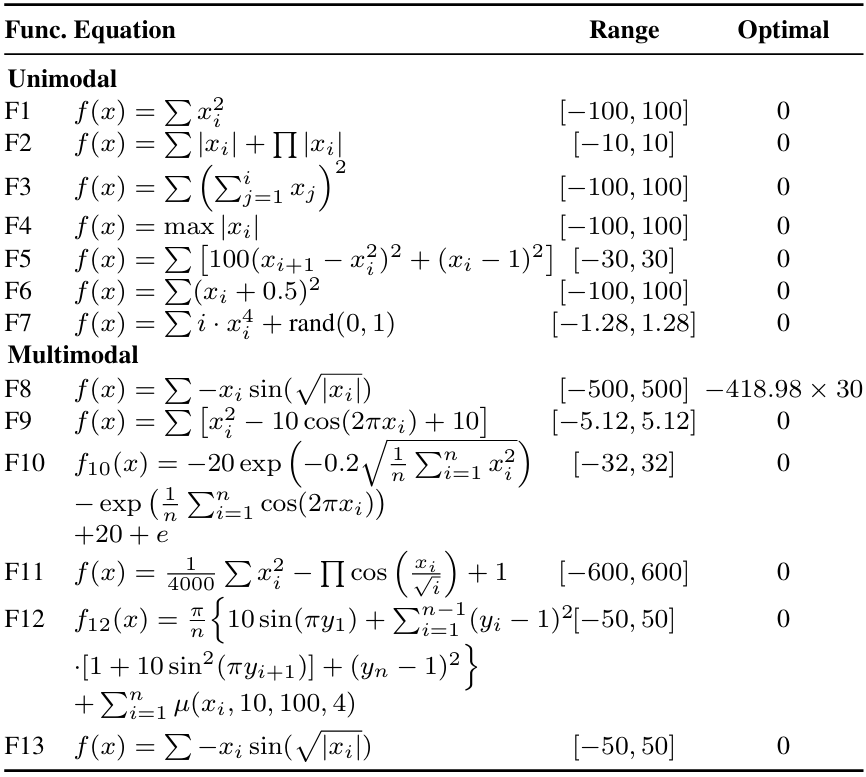} 
    \label{bench}
\end{table}

\begin{figure*}[!t]
    \centering
    \subfloat[Map 1]{\includegraphics[width=1.3in]{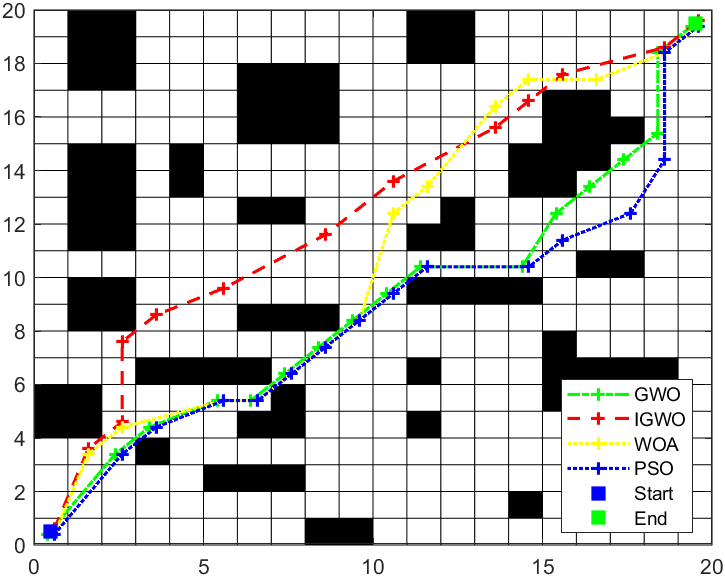}}\hspace{2pt}
    \subfloat[Map 2]{\includegraphics[width=1.3in]{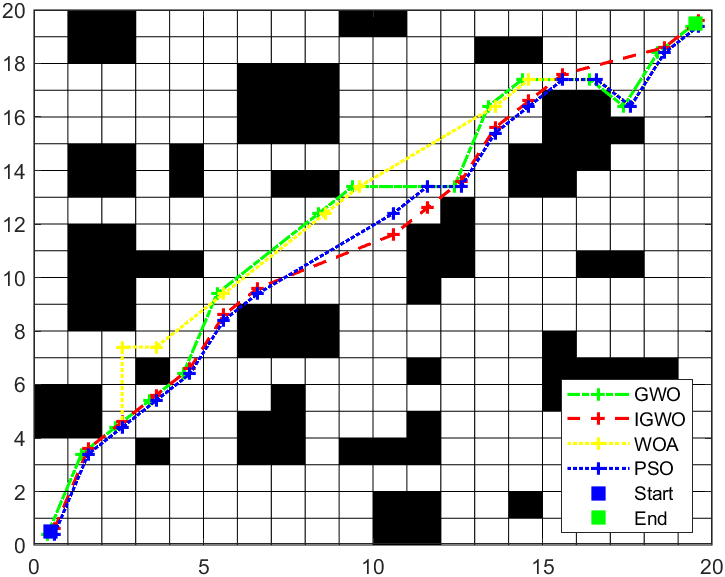}}\hspace{2pt}
    \subfloat[Map 3]{\includegraphics[width=1.3in]{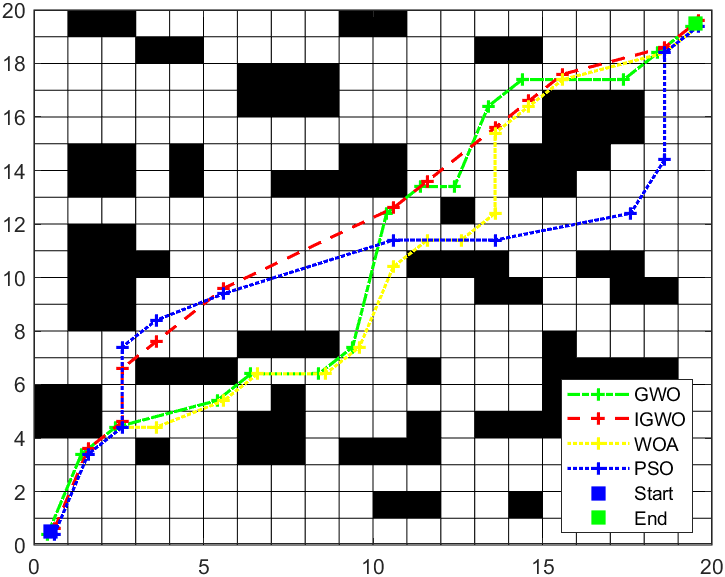}}\hspace{2pt}
    \subfloat[Map 4]{\includegraphics[width=1.3in]{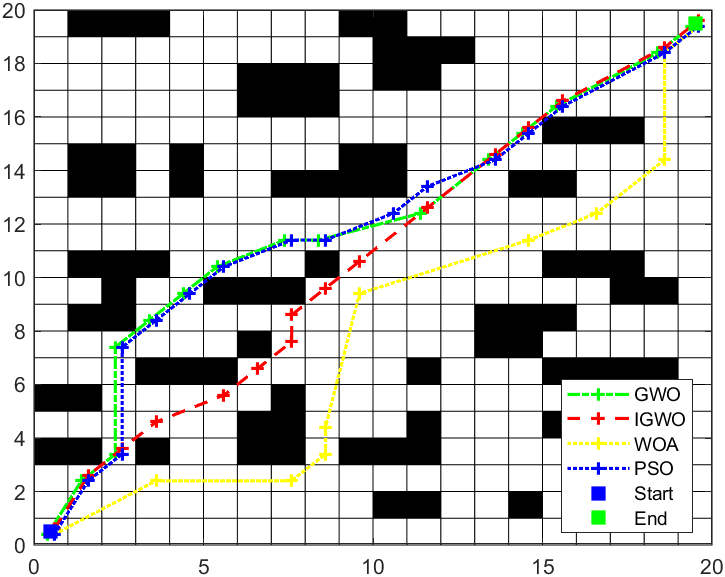}}\\[2pt]

    \subfloat[Iterative objective value changes in Map 1 scenario]{\includegraphics[width=1.3in]{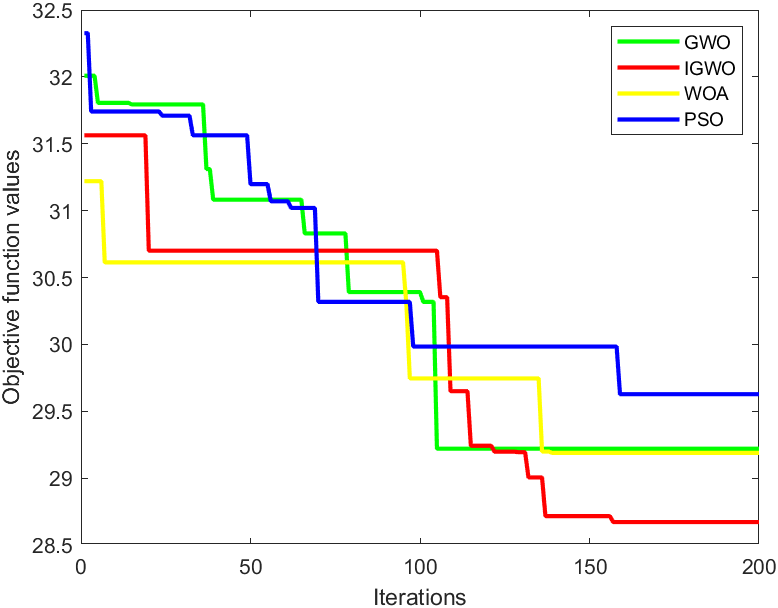}}\hspace{2pt}
    \subfloat[Iterative objective value changes in Map 2 scenario]{\includegraphics[width=1.3in]{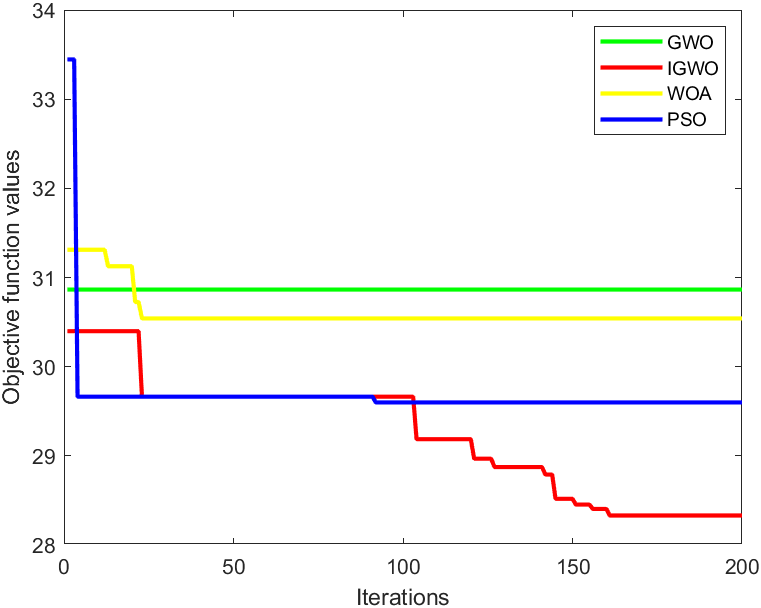}}\hspace{2pt}
    \subfloat[Iterative objective value changes in Map 3 scenario]{\includegraphics[width=1.3in]{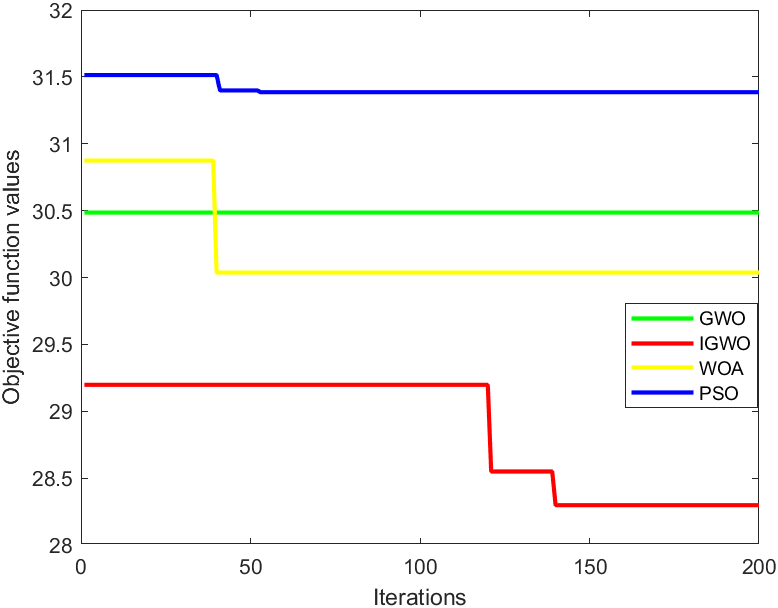}}\hspace{2pt}
    \subfloat[Iterative objective value changes in Map 4 scenario]{\includegraphics[width=1.3in]{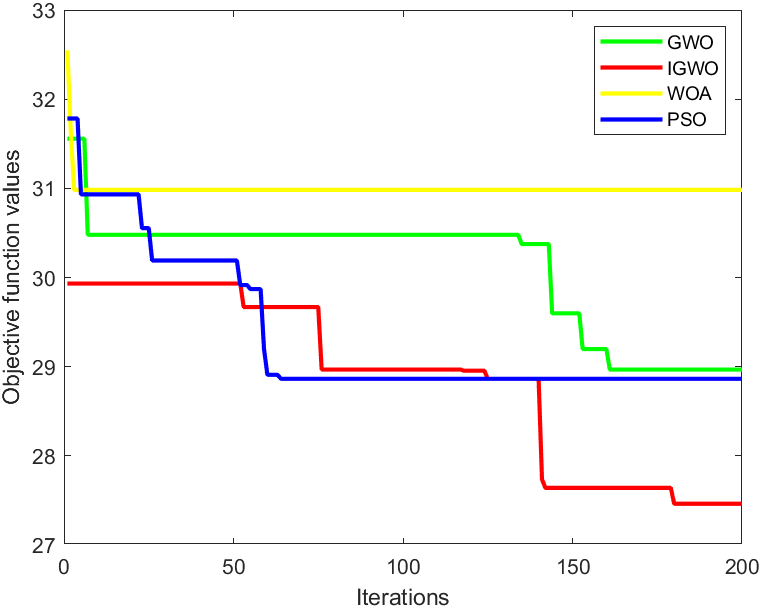}}

    \caption{The results of UAV path planning across four different map environments.}
    \label{map}
\end{figure*}

\begin{figure*}[!htbp]
\centering

\includegraphics[width=0.18\linewidth]{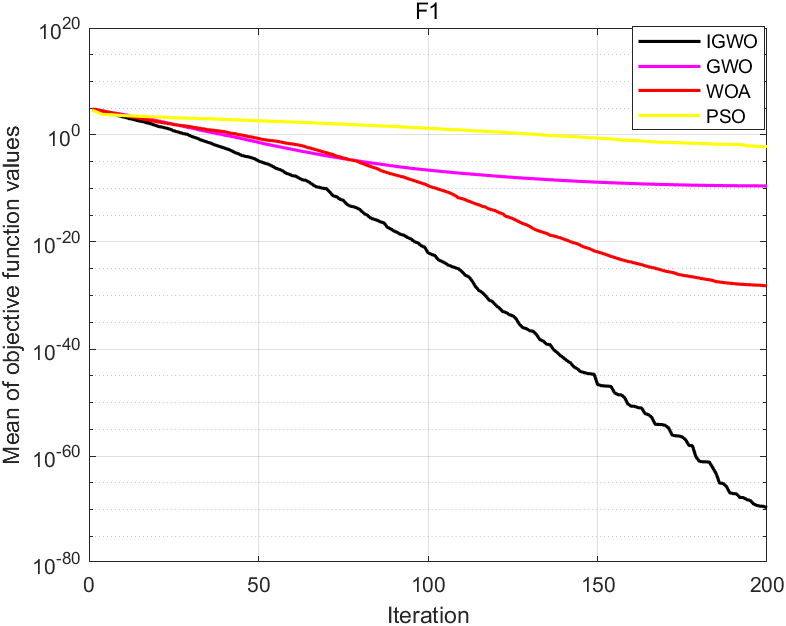}
\includegraphics[width=0.18\linewidth]{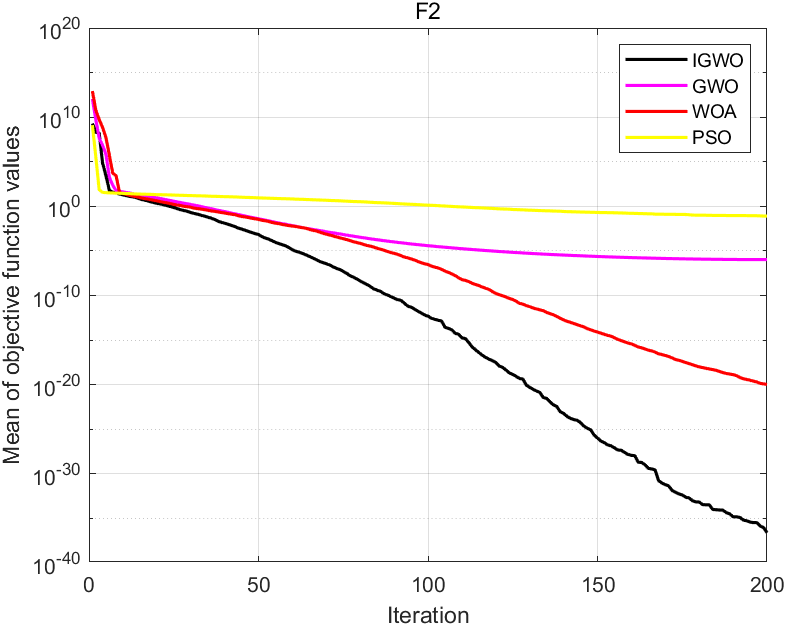}
\includegraphics[width=0.18\linewidth]{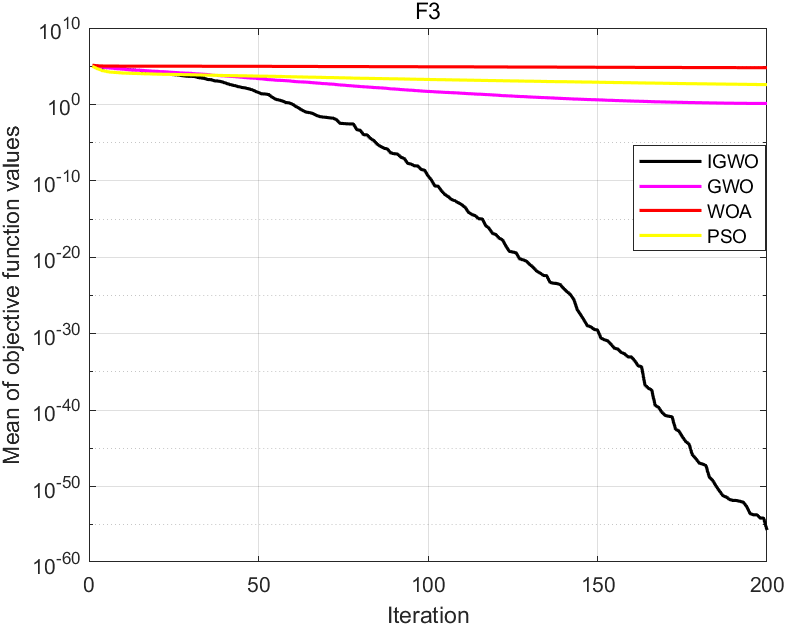}
\includegraphics[width=0.18\linewidth]{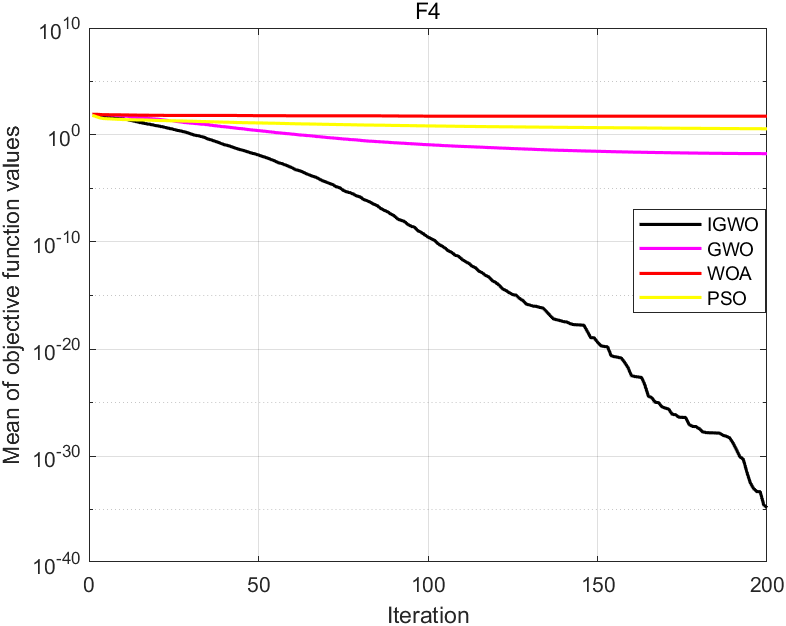}\\[2pt]

\includegraphics[width=0.18\linewidth]{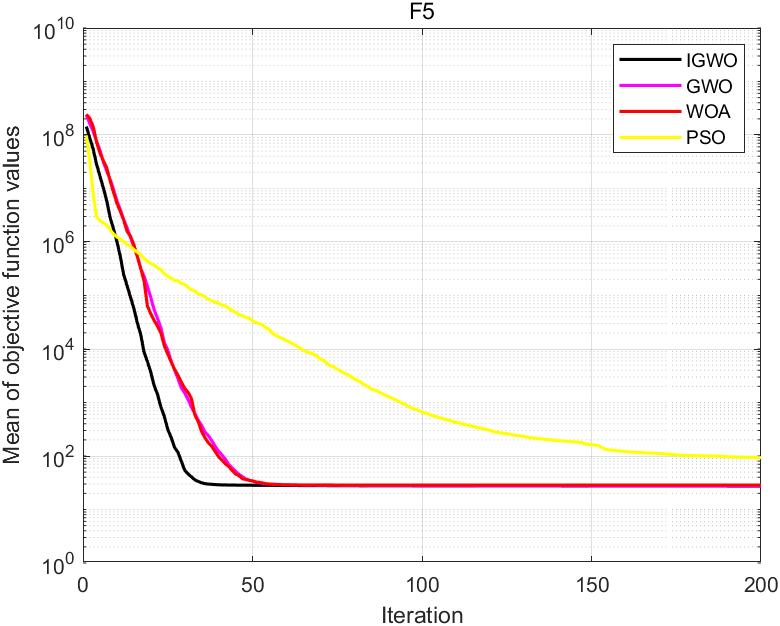}
\includegraphics[width=0.18\linewidth]{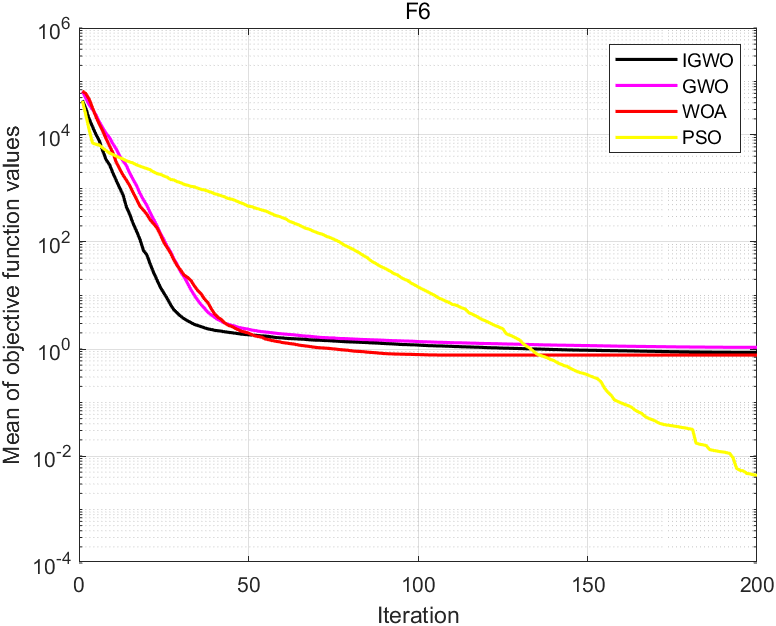}
\includegraphics[width=0.18\linewidth]{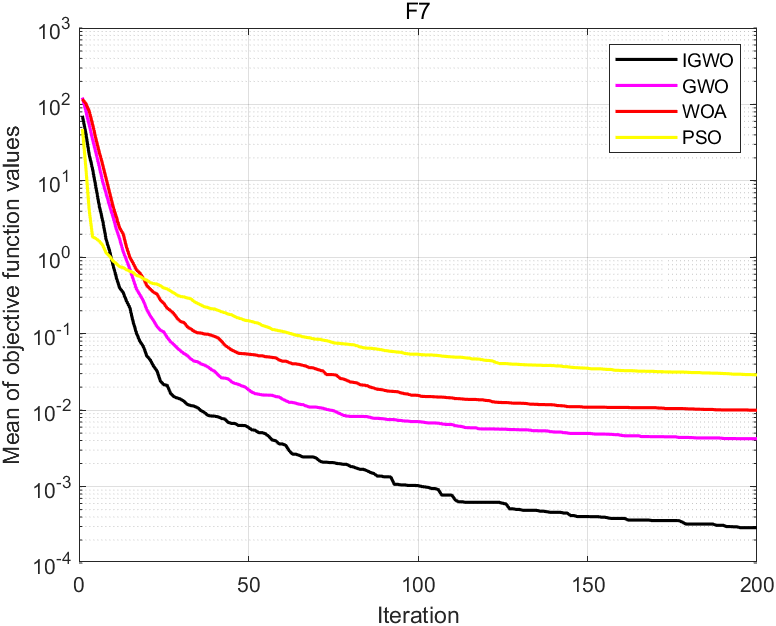}
\includegraphics[width=0.18\linewidth]{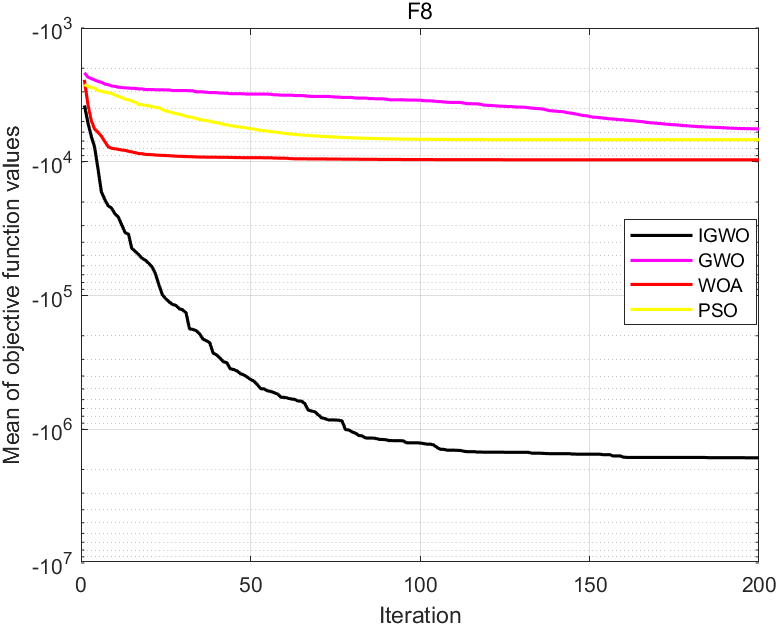}\\[2pt]

\includegraphics[width=0.18\linewidth]{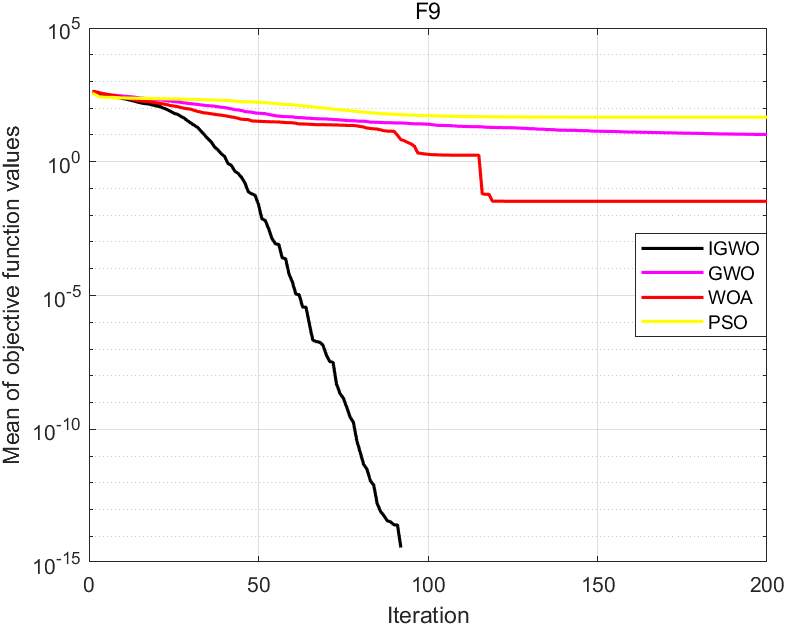}
\includegraphics[width=0.18\linewidth]{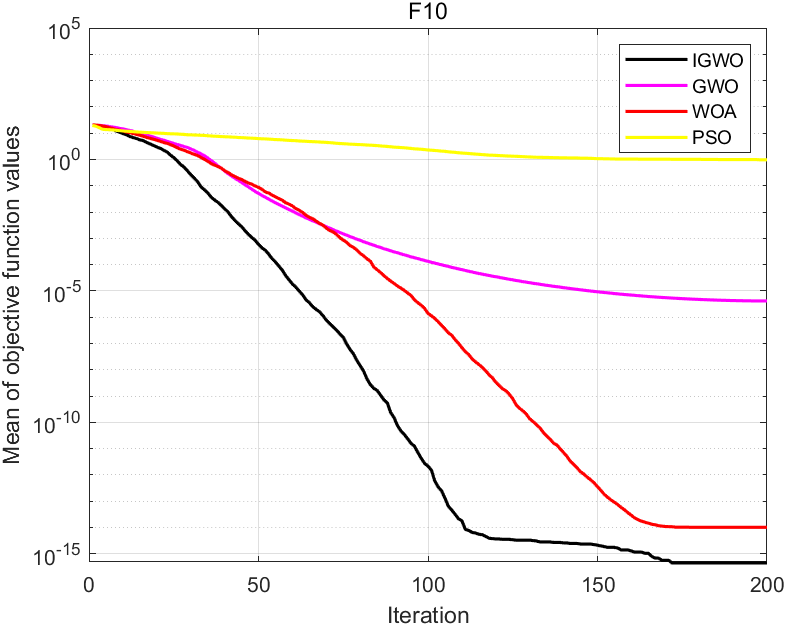}
\includegraphics[width=0.18\linewidth]{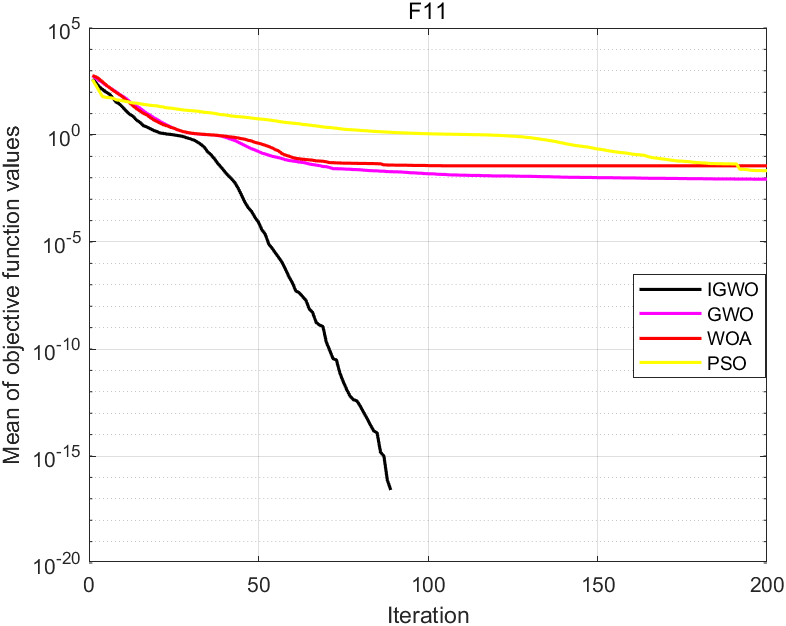}
\includegraphics[width=0.18\linewidth]{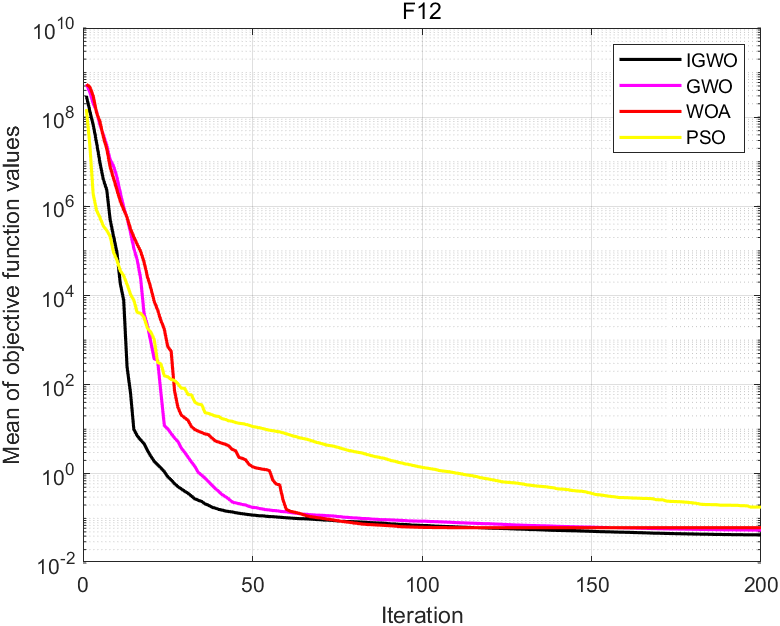}\\[2pt]

\includegraphics[width=0.18\linewidth]{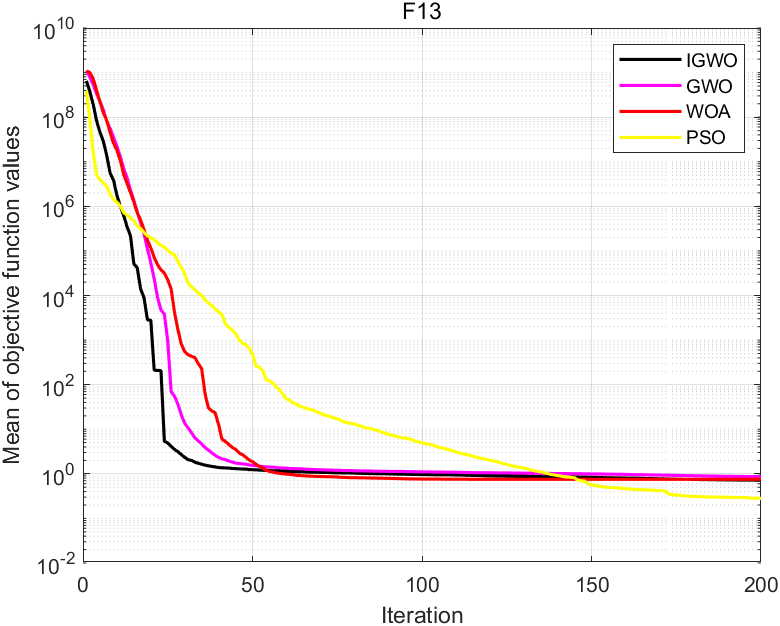}

\caption{Comparative convergence curves of IGWO, GWO, PSO and WOA on benchmark functions}
\label{IGWO_graphics1}
\end{figure*}

\section{SIMULATION RESULTS}
\subsection{Performance of IGWO on Benchmark Functions}
Heuristic algorithms are typically evaluated using a set of benchmark functions with known optimal values \cite{suganthan2005problem}. The algorithm is deemed more effective when its obtained solution closely approximates the known optimal value. The benchmark functions and their known optimal values are presented in Table \ref{bench}.

In this study, IGWO is evaluated alongside GWO, Particle Swarm Optimization (PSO) \cite{kennedy1995particle}, and Whale Optimization Algorithm (WOA) \cite{mirjalili2016whale} to assess its performance. For each benchmark function, 30 independent executions are conducted, with the resulting mean (avg) and standard deviation (std) values recorded accordingly. All compared methods operate with a swarm size of 40 and are iterated over 200 cycles. The benchmark function values for all methods are shown in Table \ref{IGWO_table}, which can be visualized in Fig. \ref{IGWO_graphics1}.

IGWO achieves the best performance on benchmark functions F1–F5, F7, and F9–F12, ranking first among all compared algorithms. Notably, it attains both an avg and std of zero on functions F9 and F11, indicating highly stable and precise optimization results. These findings demonstrate that IGWO possesses strong optimization capabilities on both unimodal and multimodal objective functions.

\begin{table}[H]
\centering
\caption{Comparison of the path lengths generated by four different methods across various map environments (unit: m)}
\label{tab:uav_maps}
\begin{tabular}{c|cccc}
\hline
\diagbox{Maps}{Methods} & IGWO & GWO & PSO & WOA \\
\hline
Map1 & 28.68 & 29.24 & 29.62 & 29.23 \\
Map2 & 28.33 & 30.87 & 29.60 & 30.54 \\
Map3 & 28.30 & 30.49 & 31.39 & 30.04 \\
Map4 & 27.46 & 28.98 & 28.86 & 30.98 \\
\hline
\end{tabular}
\end{table}

\subsection{Performance of IGWO-based UAV Path Planning Method}
In this study, the IGWO-based UAV shortest path planning method is tested in a $20\,\text{m} \times 20\,\text{m}$ grid environment and compared with the GWO, PSO, and WOA-based path planning methods. As shown in Fig. \ref{map}, black cells represent obstacles, while white cells denote navigable areas. The map is defined such that the initial position lies in its lower-left region, and the target destination is located in the upper-right. To evaluate algorithm performance, four distinct map environments are randomly generated, and IGWO is applied alongside three other heuristic algorithms to tackle the shortest route planning task. The objective function, with $m = 20$, $P = 10$, is formulated as shown in Eq. (13). The path length results produced through the application of the compared algorithms are listed in Table \ref{tab:uav_maps}, and are visually illustrated in Fig. \ref{map}.

Across all test maps, the UAV path planning method based on IGWO consistently achieves the shortest paths relative to the remaining three approaches. Specifically, in comparison to GWO, PSO, and WOA, the IGWO-based method yields average path length reductions of 1.70m, 1.68m, and 2.00m, respectively, across the four environments.

Based on the above simulation results, the IGWO-based UAV path planning method demonstrates strong robustness and superior adaptability across different environments. Compared to the other three heuristic-based methods, IGWO consistently identifies the shortest feasible path in all tested scenarios, regardless of obstacle distribution and complexity. This consistency indicates that IGWO maintains effective exploration and exploitation capabilities, even in highly constrained and dynamic environments.

\section{CONCLUSION}
In this work, a novel UAV path planning method based on IGWO is proposed. The IGWO integrates two key strategies: ACP before exploitation to intensify local search near the leader wolf, and LOBL after exploitation to enhance solution diversity via reflected agent positions inspired by convex lens imaging. These improvements strengthen both local exploitation and global exploration. On benchmark functions F1–F5, F7, and F9–F12, IGWO outperforms other algorithms and achieves the global optimum on F9 and F11. Subsequently, an IGWO-based solution is introduced to address the challenge of computing efficient UAV routes. This routing task is formulated as a search-based optimization scenario featuring a well-specified goal, wherein IGWO is utilized to search for the optimal route. Simulation results show that the paths planned by IGWO are, on average, shorter than those planned by GWO, PSO, and WOA by 1.70m, 1.68m, and 2.00m, respectively, across four different maps. Therefore, the performance advantages suggest that the proposed method is particularly suitable for mission-critical tasks such as disaster rescue, emergency response, and last-mile logistics delivery, where determining an optimal trajectory plays a key role in ensuring both operational performance and reliability.

\bibliographystyle{IEEEtran}
\bibliography{ref}

\end{document}